\documentclass[letterpaper, 10 pt, conference]{ieeeconf}  % Comment this line out if you need a4paper

\IEEEoverridecommandlockouts                              % This command is only needed if 
\usepackage[pdftex]{graphicx}
\usepackage{cite}
\usepackage{amsmath, amssymb}
\usepackage{array}
\usepackage{url}
\usepackage{xcolor}
\usepackage{blindtext}
\usepackage{booktabs}
\usepackage[section]{placeins}
\usepackage{subcaption}
\usepackage{multirow}
\usepackage{float}

\title{\LARGE \bf
ARC: Adversarially Robust Control Policies for Autonomous Vehicles
}

\author{Sampo Kuutti,
	Saber Fallah, % <-this % stops a space
	\thanks{Sampo Kuutti and Saber Fallah are with the Connected and Autonomous Vehicles Lab, University of Surrey, Guildford, GU2 7XH, UK. Email: \{s.j.kuutti, s.fallah\}@surrey.ac.uk}
	Richard Bowden
	\thanks{Richard Bowden is with the Centre for Vision, Speech and Signal Processing, University of Surrey, GU2 7XH, UK. Email: r.bowden@surrey.ac.uk}
} 

\begin{document}

\maketitle
\thispagestyle{empty}
\pagestyle{empty}

%%%%%%%%%%%%%%%%%%%%%%%%%%%%%%%%%%%%%%%%%%%%%%%%%%%%%%%%%%%%%%%%%%%%%%%%%%%%%%%%
\begin{abstract}
Deep neural networks have demonstrated their capability to learn control policies for a variety of tasks. However, these neural network-based policies have been shown to be susceptible to exploitation by adversarial agents. Therefore, there is a need to develop techniques to learn control policies that are robust against adversaries. We introduce Adversarially Robust Control (ARC), which trains the protagonist policy and the adversarial policy end-to-end on the same loss. The aim of the protagonist is to maximise this loss, whilst the adversary is attempting to minimise it. We demonstrate the proposed ARC training in a highway driving scenario, where the protagonist controls the follower vehicle whilst the adversary controls the lead vehicle. By training the protagonist against an ensemble of adversaries, it learns a significantly more robust control policy, which generalises to a variety of adversarial strategies. The approach is shown to reduce the amount of collisions against new adversaries by up to 90.25\%, compared to the original policy. Moreover, by utilising an auxiliary distillation loss, we show that the fine-tuned control policy shows no drop in performance across its original training distribution.
\end{abstract}

%%%%%%%%%%%%%%%%%%%%%%%%%%%%%%%%%%%%%%%%%%%%%%%%%%%%%%%%%%%%%%%%%%%%%%%%%%%%%%%%
\section{Introduction}
The powerful function approximation capabilities of Deep Neural Networks (DNNs) has pushed the state-of-the-art forward in multiple fields. This has lead to machine learning being adopted to learn control policies in applications such as robotic arm manipulation \cite{levine2016end, gu2017deep}, navigation \cite{zhu2017target, katyal2019uncertainty}, and autonomous driving \cite{bojarski2016end, codevilla2018end}. In recent years, there have been numerous DNN-driven approaches proposed for autonomous vehicle control, and among them Imitation Learning has attracted attention due to its ability to learn driving behaviours from human demonstration \cite{kuutti2020survey}. Imitation learning performs well in naturalistic driving and scales well to training data, but performs poorly when experiencing scenarios outside of the training distribution \cite{codevilla2019exploring, ross2011reduction}. Furthermore, these learned policies have been proven to be susceptible to attacks by adversarial agents \cite{kuutti2020training, gleave2019adversarial}. These limitations pose a challenge to adapting these learned control policies to safety-critical systems. 

We propose an adversarial learning framework, which uses imitation learning as a first training step and then improves the robustness to distribution shift by training the policy simultaneously against an ensemble of adversarial agents whose goal is to degrade the performance of the target policy. Both networks learn through a semi-competitive game, where one aims to drive in a safe manner and the other aims to create scenarios in which collisions could occur. Therefore, over time the target agent learns to avoid mistakes which an adversary could exploit. Our tests show that the approach maintains a safe behaviour even against learned adversarial agents, and results in a more robust and safe control policy.

This approach is partially inspired by the minimax game at the heart of Generative Adversarial Networks (GANs) \cite{goodfellow2014generative}, where two networks are trained on the same loss such that the Discriminator aims to correctly classify images as real or fake whilst the Generator aims to fool the Discriminator with generated images. GANs have also inspired the Generative Adversarial Imitation Learning (GAIL) \cite{ho2016generative, kuefler2017imitating}, where the Generator generates actions, whilst the Discriminator aims to predict whether the state-action pair comes from the Generator or the Expert. However, different from GANs or GAIL, where the Generator generates images/actions and the Discriminator performs binary classification, in our work both networks learn to predict continuous control actions for separate agents within a simulator. 

Image-based DNN classifiers have been shown to be susceptible to adversarial attacks, which perturb the observations of the network, causing them to misclassify the image \cite{szegedy2013intriguing, goodfellow2014explaining}. As a common defense, adversarial training has been shown to improve robustness to adversarial attack \cite{xie2019feature}. Similarly, perturbing the observations of a policy or varying its dynamics during training in an adversarial fashion has been proven to improve the robustness of learned control policies \cite{morimoto2005robust, mandlekar2017adversarially, pattanaik2018robust, pinto2017robust}. Combining concepts of competing networks from GANs and adversarial training, Robust Adversarial Reinforcement Learning (RARL) \cite{pinto2017robust, pan2019risk, ma2018improved} uses two DNNs trained through Reinforcement Learning (RL), where one DNN aims to learn a control policy for a given task and the other DNN aims to degrade the performance of the target policy by generating disturbances in its observations or actions. The RARL approach has also been shown to improve robustness of RL policies for different tasks, including autonomous driving. Going beyond disturbances in observation or action space, the Adversarial Policies approach by Gleave \textit{et al.} \cite{gleave2019adversarial} controls a separate agent in the same environment as the target policy, where the adversarial agent aims to prevent the target agent from performing in its task successfully. The Adversarial Policy was shown to learn behaviours which significantly weaken the performance of the target policy, and by fine-tuning the target policy through RL, it learned to counter the adversary. However, it was shown that new adversaries could be trained to find new weaknesses even in the fine-tuned target policy. Concurrently, several approaches have emerged in autonomous vehicle testing, where an adversarial policy is used to control an agent (e.g. vehicle, pedestrian) on the road, and aim to find behaviours which cause the target autonomous vehicle to make mistakes \cite{koren2018adaptive, kuutti2020training, behzadan2019adversarial, ding2020multimodal}. This type of adversarial testing has been shown to be effective in the validation of autonomous vehicle control policies, by finding weaknesses which may not have been found through traditional validation methods \cite{corso2020survey, riedmaier2020survey}. 

In this work we utilise similar adversarial agents to exploit weaknesses in the target control policy, but rather than training each agent independently, we employ a GAN-like minimax loss where the agents are trained end-to-end to compete against each other. This results in more robust control policies. We show that by taking an initially susceptible Imitation Learning vehicle motion control policy, and fine-tuning it through our ARC training framework, the policy learns to avoid collisions against the competing adversary. Moreover, we show that after adversarial fine-tuning, the resulting control policy exhibits significantly improved robustness to new adversarial agents trained against it. We also demonstrate that using an auxiliary distillation loss results in the fine-tuned control policy retaining the same level of performance across its original training distribution, thereby improving robustness to safety-critical scenarios without degrading performance in typical driving scenarios.

The remainder of this paper is as follows. Section II describes the methodology used for pre-training of the target and adversary control policies, as well as the proposed Adversarially Robust Control framework for training both networks end-to-end. The simulated experimental results are presented and discussed in Section III. Finally, concluding remarks are given in Section IV.

%%%%%%%%%%%%%%%%%%%%%%%%%%%%%%%%%%%%%%%%%%%%%%%%%%%%%%%%%%%%%%%%%%%%%%%%%%%%%%%%
\section{Methodology}
We demonstrate our approach in a vehicle following scenario applied to highway driving. The aim of the host vehicle is to maintain a safe distance from the lead vehicle in front. To do this, the control policy infers actions which control the gas and brake pedals of the host vehicle, based on the low-dimensional states from the vehicle's radar and inertial sensors. The adversarial agent controls the lead vehicle, and is trained through Reinforcement Learning to create scenarios in which collisions are likely to occur.  We first describe the training methodology for the Imitation Learning (IL) based host vehicle control policy, followed by the training of the adversarial agent. Finally, we describe our Adversarially Robust Control (ARC) formulation, where both agents are trained end-to-end through a minimax loss. We denote the Imitation Learning based agent by $IL$, while during the ARC training, where both networks are trained end-to-end, the Protagonist and Adversary are denoted by $P$ and $A$, respectively.

\subsection{Imitation Learning}
Imitation Learning is a subset of Supervised Learning, where the model learns from expert demonstrations of trajectories \cite{pomerleau1991efficient}. Imitation learning aims to learn a control policy by imitating the behaviour of an expert, by observing states $s^{IL}_t$ and predicting a corresponding control action $a^{IL}_t$, which is then compared to the expert's optimal action $\hat{a}_t$. This can be done by collecting a dataset $\mathcal{D} =\{s_t, \hat{a}_t\}^{N}_{t=0}$ of $N$ expert demonstrations, and then training the agent to predict the expert's actions for the states in the dataset in a supervised manner. In this work, we use the Imitation Learning based vehicle motion control model from \cite{kuutti2019safe}, which trains a feedforward neural network through Imitation Learning to predict the longitudinal control actions of a vehicle in highway driving. The Imitation Learning policy is denoted by $\pi^{IL}$ and is represented by a feedforward neural network with 3 hidden layers of 50 neurons each, with the parameters $\theta^{IL}$. Therefore, the agent's aim is to learn a policy $\pi^{IL}$ which generates actions similar to the expert policy $\pi^*$, by finding the optimal parameters $\theta^*$ based on an imitation loss $\mathcal{L}^{IL}$:
\begin{equation}
\theta^* = \arg \min_{\theta^{IL}} \sum_{t}\mathcal{L}^{IL}(\pi^{IL}(s^{IL}_t|\theta^{IL}), \hat{a}_t)
\end{equation}
The network is trained using the Mean Square Error (MSE) loss with respect to the labels given by the expert's action in dataset $\mathcal{D}$:
\begin{equation}
\mathcal{L}^{IL}(\mathcal{D}, \pi^{IL}(s^{IL}_t|\theta^{IL}), \hat{a}_t) = | a^{IL}_t - \hat{a}_t | ^ 2
\end{equation}

The dataset was collected by driving at highway speeds on a single road, within the IPG CarMaker simulator \cite{IPG2017}. The expert demonstrator used to collect example actions, is the default driver in the simulator, IPG Driver. The expert's aim is to maintain a 2s time headway, $t_h$, from the lead vehicle in front of the host vehicle. The time headway $t_h$ is a measure of distance between two vehicles in time, as given by:
\begin{equation}
t_h = \frac{x_{rel}}{v}
\end{equation}
where $x_{rel}$ is the distance between the two vehicles in m, and $v$ is the velocity of the host vehicle in m/s.

Each observation $s_t$ in the dataset consists of the host vehicle velocity $v$, relative velocity with respect to the lead vehicle $v_{rel}$, and time headway $t_h$, such that $s_t^{IL} = (v, v_{rel}, t_h)$. The action of the agent controls the vehicle's gas and brake pedals, and is represented as a single continuous value $a^{IL}_t \in [-1, 1]$, where negative values represent the use of the brake pedal and positive values represent the use of the gas pedal. 

\subsection{Adversarial Reinforcement Learning}
Reinforcement learning can be formally described by a Markov Decision Process (MDP) denoted by a tuple ($\mathcal{S}$, $\mathcal{A}$, $\mathcal{P}$, $\mathcal{R}$, $\gamma$), where $\mathcal{S}$ is the state-space, $\mathcal{A}$ is the action-space, $\mathcal{P}$ is the transition probability model, $\mathcal{R}$ is the reward function, and $\gamma$ is the discount factor. At every timestep $t$, the RL agent observes the state $s_t \in \mathcal{S}$ and takes an action $a_t \in \mathcal{A}$ according to its policy $\pi$. Then, the environment $E$ transitions to the next state $s_{t+1}$ according to the state transitions probability $p(s_{t+1}|s_t, a_t)$ as given by $\mathcal{P}$. The agent then receives a scalar reward $r_t \in \mathcal{R}$. The aim of the RL agent is to maximise its long term discounted rewards, as given by the returns $R_t$:
\begin{equation}
R_t = \sum_{k=0}^{\infty}\gamma^kr_{t+k}
\end{equation}
where the discount factor $\gamma \in [0, 1]$ is used to prioritise immediate rewards over future rewards.

To find weaknesses in the target control policy, we employ the Adversarial Testing Framework by Kuutti \textit{et al.} \cite{kuutti2020training} based on Adversarial Reinforcement Learning (ARL). The technique uses an agent trained through reinforcement learning, whose aim is to create collisions with the vehicle behind it. Therefore, this agent acts as a lead vehicle to the host vehicle control policy described in the previous subsection. However, to ensure the results are realistic and all collisions are preventable (and therefore any collisions mean the host vehicle made mistakes), the actions and states of the adversarial agents are constrained. In \cite{kuutti2020training}, the robustness of vehicle follower policies were tested in different velocity ranges, and the velocity range with the most collisions was $v_{lead} \in [12, 30]$ m/s. Therefore, we utilise these velocity limits for the adversary, and aim to reduce collisions by improving the robustness of the protagonist, whilst minimising any impact on the agent's behaviour in its training domain. Similarly, to ensure the collisions are avoidable, the acceleration of the lead vehicle is limited to $a_{lead} \in [-6, 2]$ m/s\textsuperscript{2}. During training of the adversarial agent, various values for the coefficient of friction in the ranges [0.4, 1.0] were used to test the response of the target agent in different driving conditions. The adversary's observations are represented by $s^A_t = (v, a, v_{rel}, t_h)$, where $v$ is the velocity and $a$ is the acceleration of the following vehicle. The action of the adversary $a^A$ is a continuous value for the acceleration of the lead vehicle $a_{lead}$. The adversarial agent is trained through Advantage Actor Critic (A2C) \cite{mnih2016asynchronous} Reinforcement Learning, which is an actor-critic on-policy algorithm. The two networks, actor and critic networks, estimate the policy function $\pi^{A}$ and value function $V(s^A)$. To improve training stability, the weights of both networks are updated based on the Advantage function $A(s^A, a^A)$:
\begin{equation}
V(s^A) = \mathbb{E}[R_t|s^A_t = s^A]
\end{equation}
\begin{equation}
Q(s^A, a^A) = \mathbb{E}[R_t|s^A_t = s^A, a^A]
\end{equation}
\begin{multline}
A(s^A_t, a^A_t) = Q(s^A_t, a^A_t) - V (s^A_t) \\
\approx \sum_{k}^{n-1} \gamma^kr_{t+k} + \gamma^n V(s^A_{t+n}) - V(s^A_t)
\end{multline}
Where $\mathbb{E}$ denotes expectation, $V(s^A_t)$ is the value function, and $Q(s^A_t, a^A_t)$ is the state-action (or quality) function \cite{sutton1998reinforcement}.

To estimate the stochastic policy $\pi^A$, the actor network uses two outputs, estimated action value $\mu$ and estimated action variance $\sigma^2$. The action applied by the adversarial agent is then sampled from the Gaussian distribution $a^A_t\sim\mathcal{N}(\mu, \sigma^2)$. To do this, the actor network uses 3 hidden layers with 50 neurons, followed by a Long Short-Term Memory \cite{hochreiter1997long} layer with 16 units, followed by the output layer. Meanwhile, the critic network estimating the value function $V(s^A)$, uses 2 hidden layers with 50 neurons. All hidden neurons use the ReLU-6 activation, $\mu$ uses a tanh activation, $\sigma^2$ uses a softplus activation, and the value estimate uses a linear activation. To train both networks, A2C updates the actor network parameters $\theta^{\pi}$ and critic network parameters $\theta^{V}$, using the policy loss $\mathcal{L}_{\pi^A}$ and value loss $\mathcal{L}_V$, respectively:
\begin{equation}
\mathcal{L}_{v} = (A(s^A_t, a^A_t))^2
\end{equation}
\begin{equation}
\mathcal{L}_{\pi^A} = -log\pi^A(a^A_t| s^A_t)A(s^A_t,a^A_t) - \beta H(\pi^A(s^A_t))
\end{equation}
where $\beta$ is the entropy coefficient and $H(\pi^A(s^A_t))$ is the policy entropy used to encourage exploration in the adversary's policy, given by
\begin{equation}
H(\pi^A(s^A_t)) = \frac{1}{2}(log(2\pi \sigma^2)+1)
\end{equation}
Both networks were trained using the RMSProp optimiser \cite{tieleman2012lecture}, using their respective losses.

To train the adversarial agent to find collisions against target policies it was trained using the adversarial reward function based on inverse headway given by:
\begin{equation}
r^A(s^A, a^A) =  \min{\left(\frac{1}{t_h}, 100\right)}
\end{equation}
where $r^A$ is the adversary's reward, and the reward is capped at 100 to avoid the reward tending towards infinity as the headway approaches zero.

%%%%%%%%%%%%%%%%%%%%%%%%%%%%%%%%%%%%%%%%%%%%%%%%%%%%%%%%%%%%%%%%%%%%%%%%%%%%%%%%
\subsection{Adversarially Robust Control (ARC)}
The Adversarially Robust Control framework utilises two networks, the Protagonist network $P$ and the Adversary network $A$, initialised from the IL network (Section II-A) and ARL network (Section II-B), respectively. The scenario where both networks are learning to compete against each other can be formulated as a two player Markov Game, which is a multi-agent game theoretic formulation of an MDP \cite{littman1994markov, perolat2015approximate}. The Markov Game can be strictly competitive (zero-sum) or semi-competitive (nonzero-sum), depending on whether the agents are directly competing against each other or whether they have additional objectives \cite{ma2018improved}. The Markov Game with Protagonist $P$ and Adversary $A$ is denoted by a tuple ($\mathcal{S}$, $\mathcal{A}^A$, $\mathcal{A}^P$, $\mathcal{P}$, $\mathcal{R}^A$, $\gamma$). The $P$ and $A$ observe states $s_t^P \in \mathcal{S}$ and $s_t^A \in \mathcal{S}$ and take actions $a_t^P \in \mathcal{A}^P$ and $a_t^A \in \mathcal{A}^A$, respectively. The environment $E$ then transitions to the next state according to transition model $\mathcal{P}$, and the adversary receives a reward $r_t^A \in \mathcal{R}^A$. Note, unlike RARL approaches with two RL agents, we do not define a reward for the Protagonist, rather the $P$ network directly maximises the policy loss of the adversary, such that both agents are trained end-to-end using the same loss:
\begin{multline}
\min_{A}\max_{P}\mathcal{L}_{\pi^A}(A, P) \\
= -log\pi^A(a^A_t| s^A_t)A(s^A_t,a^A_t) - \beta H(\pi^A(s^A_t))
\end{multline}
Therefore, the aim of the Adversary is to maximise its reward function $r^A$, which encourages the agent to take actions which lead the following vehicle to collide into it. Meanwhile, the Protagonist aims to maximise this loss, effectively aiming to take actions which lead to lower rewards for the adversary, and thus less collisions. Having the $P$ network directly maximise the Adversary's policy loss has the advantage that no additional training signal has to be engineered for the Protagonist (e.g. labels for supervised learning or rewards for reinforcement learning). This also makes the proposed framework more general, as it is agnostic to the learning technique used for pre-training (e.g. no assumptions about the stochasticity of the policy) and simply needs access to the weights of the $P$ network. The Adversary used here differs from the one in Section II-B, in that it uses an additional observation, which is the action taken by the protagonist $a^P$. Therefore $s_t^A = (v, a, v_{rel}, t_h, a^P)$, making the output of the $A$ network a function of the $P$ network; $a^A = A(s^A) = A(P(s^P))$, and the policy loss $\mathcal{L}_{\pi^A}$ is differentiable with respect to both $P$ and $A$. We train both networks in the highway driving scenario where the Protagonist controls the follower vehicle, whilst the Adversary controls the lead vehicle. Each training episode lasts for 5 minutes or until a collision occurs. The training is sped up by using the DNN-based simulator proxy described in \cite{kuutti2019end}, which acts as a type of World Model \cite{ha2018world} estimating the simulator, and was shown to speed up training by up to a factor of 20. Further testing is later carried out in IPG CarMaker simulator to validate the control policy performance (Section III).

However, while naively maximising the policy loss in a strictly competitive game setting would lead to behaviours which degrade the performance of the adversary, it does not necessarily provide robust policies which generalise to different lead vehicle behaviours. We show that this type of competitive game setting causes the $P$ agent to either learn an overly conservative driving strategy or to overfit to the adversarial lead vehicle while forgetting how to drive in non-adversarial scenarios. Instead, we propose to use a semi-competitive game setting where an auxiliary loss is used for training the $P$ network, ensuring it does not overfit to the adversarial scenarios or catastrophically forget how to perform in its original state distribution.

The first possible issue with learning only from the adversary is becoming overly conservative to avoid collisions or overfitting to the adversarial scenarios created by the adversary. Since such driving scenarios represent edge-cases, which during normal driving would only occur rarely, there is the potential risk for the Protagonist to forget how to perform well in the natural driving scenarios. This is a similar issue to the catastrophic forgetting \cite{french1999catastrophic, goodfellow2013empirical}, which can occur in domain adaption when the model adapts to a new domain and forgets the previous domain \cite{li2017learning, jung2017less}. Indeed, in Adversarial Policies, Gleave \textit{et al.} \cite{gleave2019adversarial} noted that fine-tuning target policies against adversaries leads RL policies to forget how to perform against normal opponents. Therefore, to avoid overfitting the $P$ network to the adversarial scenarios, an auxiliary distillation loss $\mathcal{L}_D$ is defined which discourages the network from changing its behaviour drastically from the un-tuned IL model. This concept is similar to knowledge distillation \cite{hinton2015distilling} or policy distillation \cite{rusu2015policy}, however here the distillation loss is used to prevent catastrophic forgetting when training in a new distribution instead of distilling the policy into a smaller network. The $\mathcal{L}_D$ loss uses supervision from the un-tuned IL network by penalising the actions of the $P$ model based on the absolute difference to the action which would have been taken by the original IL model for the same state:
\begin{equation}
\mathcal{L}_D = \Vert a^P - a^{IL} \Vert
\end{equation}

Such that the final loss minimised by the Protagonist becomes:
\begin{equation}
\mathcal{L}_P = -\mathcal{L}_{\pi^A} + \lambda \mathcal{L}_D
\end{equation}
where $\lambda$ is a scaling hyperparameter.

\begin{figure}
	\centering
	\includegraphics[width=0.5\textwidth]{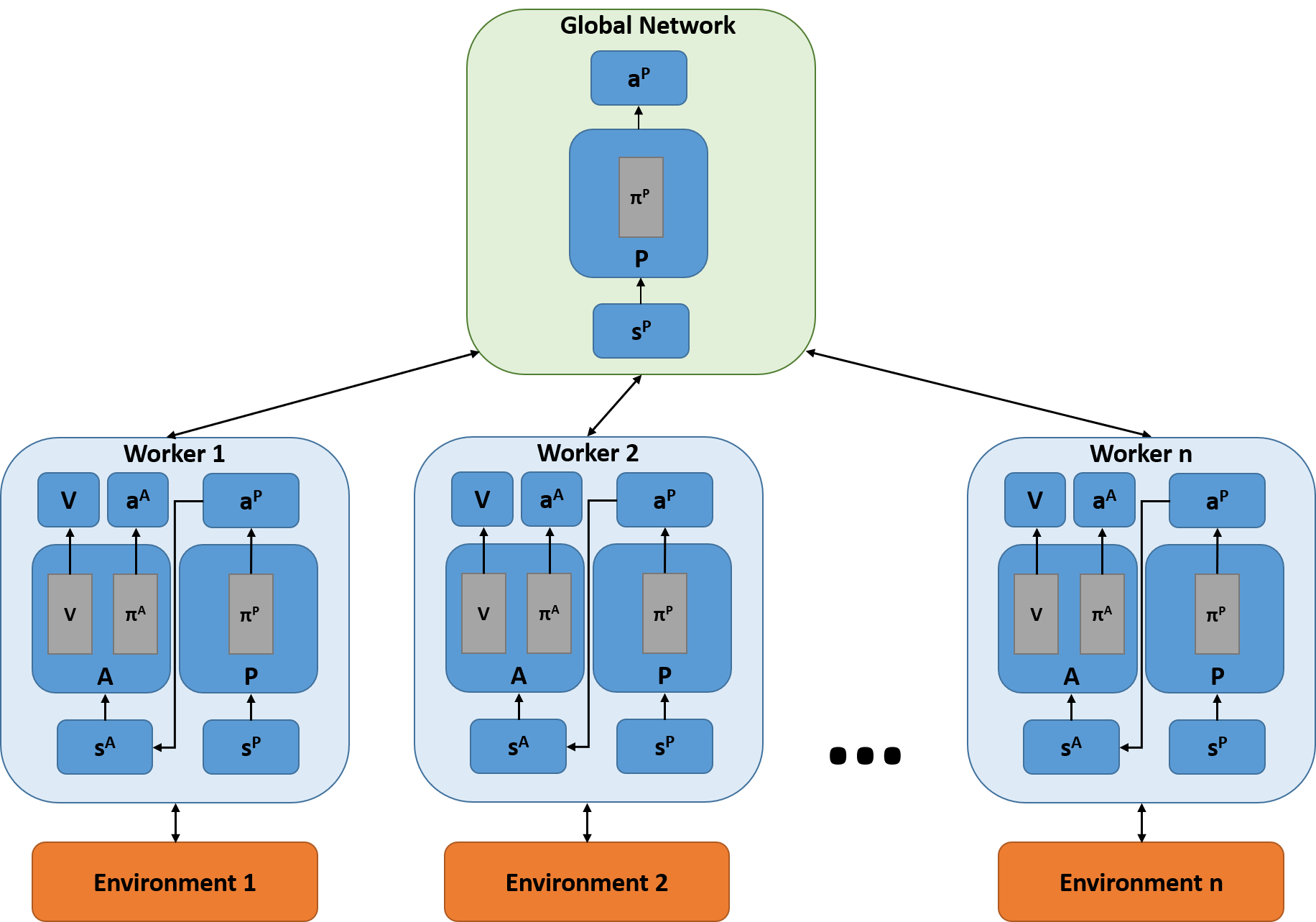}
	\caption{Training environment.}
	\label{fig_env}
	
\end{figure}

A second possible overfitting issue with this framework is overfitting due to repetitive similar behaviour of the Adversary. Different from Adversarial Policies \cite{gleave2019adversarial}, which fine-tuned against fixed adversarial policies, we train both the $P$ and $A$ simultaneously, allowing the $A$ to adapt as the $P$ learns to counter it. However, this alone may not be enough, as the $A$ network may get stuck in a local minima and continue to use the same strategy or it may adapt slowly to the improved robustness of the $P$ network. Therefore, we train the $P$ network in multiple environments simultaneously, where each environment $E_i$ uses a different adversary $A_i$, where $i = \{1, 2, .. , n\}$ for $n$ total environments. The network updates calculated based on these environments are done asynchronously, using the Asynchronous Advantage Actor Critic (A3C) \cite{mnih2016asynchronous} formulation, where each instance of the simulation $E_i$ copies the parameters of the global network to its own local network, where gradients are computed based on the experiences collected in $E_i$ by the local network. The gradients are then used to update the global network, and the local network copies the new parameters from the global network. However, in our formulation the adversaries are different agents with different parameters $\theta ^{A_i}$, therefore the global network tracks the parameters of the $P$ network, while each adversary $A_i$ is updated in the local network only, as shown in Fig. \ref{fig_env}. The Adversaries adapt to try to beat the Protagonist independently, allowing them to explore and learn different strategies, whilst the Protagonist is optimised against all Adversaries asynchronously.

%%%%%%%%%%%%%%%%%%%%%%%%%%%%%%%%%%%%%%%%%%%%%%%%%%%%%%%%%%%%%%%%%%%%%%%%%%%%%%%%
\section{Results}
Using the described formulation, we pre-train 5 adversarial agents against the IL model for 2500 episodes. Then, we train the $P$ and $A_i$ networks end-to-end for 2500 episodes, experimenting with different number of environments $n = \{1, 5, 10, 25, 50\}$. The training hyperparameters can be found in Table \ref{tbl_hyper}. As an ablation study, we also train 2 baselines to investigate the benefits of the suggested framework; ARC with a fixed single adversary and no $\mathcal{L}_D$ loss (ARC Adv. fixed, $\lambda = 0$) and ARC with a fixed single adversary (ARC Adv. fixed). We evaluate the performance during training, as well as the final trained control policies under two testing frameworks. Naturalistic driving tests the models in driving scenarios similar to those seen during training, and tests whether tuning the models against adversaries has degraded their performance in the original training distribution. The adversarial testing trains new adversaries against the control policy, and provides a measure of robustness against adversarial agents.

\subsection{Training}
\begin{figure}
	\centering
	\includegraphics[width=0.45\textwidth]{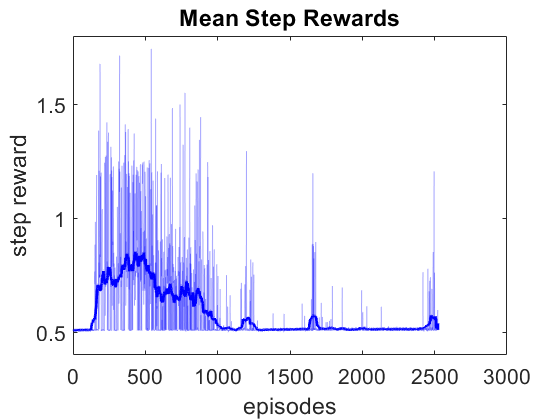}
	\caption{Mean step rewards for the adversary during ARC training. The plot shows the running mean reward (with window size of 50), with the true rewards in the transparent plot.}
	\label{fig_training}
	
\end{figure}

The training results are shown in Fig. \ref{fig_training}, where the mean step adversary rewards are visualised. Note, we show mean step reward instead of episode rewards/returns, as episodes with collisions can have significantly lower episode rewards as there are less steps to accumulate rewards. However, an episode with a collision is a successful episode for the adversary, and the higher mean step reward in such episodes reflects that. The  rewards shown in Fig. \ref{fig_training} plot the performance during training with $n = 25$. It can be seen that the Adversary initially improves its performance against the Protagonist, with increasing step rewards in the first 1000 episodes. However, over the training process, the Protagonist becomes more robust, and the mean step rewards converge $r^A_t \approx 0.5$, which corresponds to a headway of 2s.

\begin{table}
	\caption{ARC training parameters.}
	\label{tbl_hyper}
	\centering
	\begin{tabular}{ m{5cm}  m{1cm} }
		\hline
		\hline
		\textbf{Parameter} & \textbf{Value} \\
		\hline
		Adversary learning rate (actor), \textit{$\eta_{actor}$} & 1x10\textsuperscript{-4} \\
		Adversary learning rate (critic), \textit{$\eta_{critic}$} & 1x10\textsuperscript{-2} \\
		Protagonist learning rate, \textit{$\eta_{P}$} & 1x10\textsuperscript{-5} \\
		Scaling parameter, \textit{$\lambda$} & 5x10\textsuperscript{4} \\
		Discount factor, \textit{$\gamma$} & 0.99 \\
		Entropy coefficient, \textit{$\beta$} & 1x10\textsuperscript{-4} \\
		\hline
		\hline
	\end{tabular}

\end{table}

\subsection{Validation}
To understand the final performance of the fine-tuned policy, we employ two testing strategies for different driving conditions; \emph{naturalistic driving} tests the control policy in typical driving conditions similar to those seen during imitation learning, and \emph{adversarial testing} trains 5 new adversaries against the control policy and validates the robustness of the fine-tuned policy against adversarial agents and safety-critical edge case scenarios. The naturalistic testing is carried out in IPG CarMaker with different highway driving scenarios with lead vehicle velocities in the range [17, 40] m/s, acceleration [-6, 2] m/s\textsuperscript{2}, and road friction coefficient in [0.4, 1.0]. The adversarial testing trains 5 new agents against the target policy for 2500 episodes as described in Section II-B.

\begin{table*}
	\renewcommand{\arraystretch}{1.2}
	\caption{Testing of final control policies under Natural (Nat.) and Adversarial (Adv.) Testing frameworks, with baseline comparison including Imitation Learning and different versions of Adversarially Robust Control.}
	\label{tbl_simresults}
	\centering
	\resizebox{1.0\linewidth}{!}{
		\begin{tabular}{c c c c c c c c c c }
			\hline
			\hline
			\textbf{Testing Framework} & \textbf{Parameter} & 
			\begin{tabular}{@{}c@{}}\textbf{IL} \\  \scriptsize\textbf{\cite{kuutti2019safe}} \end{tabular}  & 
			\begin{tabular}{@{}c@{}}\textbf{ARC} \\  \scriptsize\textbf{Adv. fixed, $\lambda = 0$} \end{tabular} & 
			\begin{tabular}{@{}c@{}}\textbf{ARC} \\  \scriptsize\textbf{Adv. fixed} \end{tabular} & 
			\begin{tabular}{@{}c@{}}\textbf{ARC} \\  \scriptsize\textbf{n = 1} \end{tabular} & 
			\begin{tabular}{@{}c@{}}\textbf{ARC} \\  \scriptsize\textbf{n = 5} \end{tabular} & 
			\begin{tabular}{@{}c@{}}\textbf{ARC} \\  \scriptsize\textbf{n = 10} \end{tabular} & 
			\begin{tabular}{@{}c@{}}\textbf{ARC} \\  \scriptsize\textbf{n = 25} \end{tabular} & 
			\begin{tabular}{@{}c@{}}\textbf{ARC} \\  \scriptsize\textbf{n = 50} \end{tabular} \\
			\hline
			\multirow{7}{*}{Nat. Testing} 
			& min. x\textsubscript{rel} [m] & 23.84 & 49.95 & 0.00 & 32.25 & 23.66 & 23.61 & 23.61 & 23.60 \\
			& mean x\textsubscript{rel} [m] & 57.37 & 584.76 & 81.81 & 59.78 & 57.35 & 57.35 & 57.35 & 57.36 \\
			& max. v\textsubscript{rel} [m/s] & 8.88 & 15.86 & 35.54 & 3.15 & 8.92 & 9.00 & 9.02 & 9.02 \\
			& mean v\textsubscript{rel} [m/s] & 0.0197 & 2.1350 & 0.0828 & 0.0368 & 0.0217 & 0.0205 & 0.0207 & 0.0211 \\
			& min. t\textsubscript{h} [s] & 1.74 & 1.97 & 0.00 & 1.55 & 1.74 & 1.74 & 1.74 & 1.74 \\
			& mean t\textsubscript{h} [s] & 1.99 & 21.08 & 3.30 & 2.02 & 1.99 & 1.99 & 1.99 & 1.99 \\
			& collisions & 0 & 0 & 55 & 0 & 0 & 0 & 0 & 0 \\
			\hline
			\multirow{2}{*}{Adv. Testing}
			& collisions against adversaries & 800 & 0 & 2490 & 1150 & 456 & 224 & 78 & 320 \\ 
			& episodes until collision & 245 & - & 3 & 16 & 538 & 532 & 1146 & 775 \\
			\hline
			\hline
		\end{tabular}
	}
	
\end{table*}

\begin{figure}
	\centering
	\includegraphics[width=0.45\textwidth]{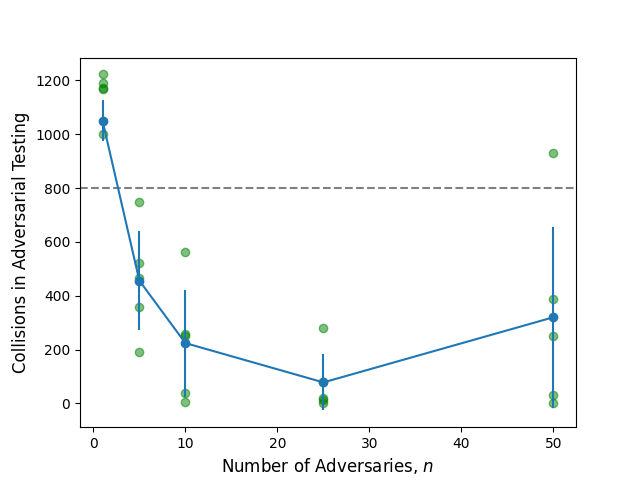}
	\caption{Collisions for different number of adversaries during Adversarial Testing. Averaged over 5 training runs, individual collision numbers visualised by green markers, mean collisions by blue markers, and standard deviation by the error bars. The dashed line indicates the level of performance by the IL model before fine-tuning.}
	\label{fig_collsperadv}

\end{figure}

\begin{figure}
	\centering
	\includegraphics[width=0.45\textwidth]{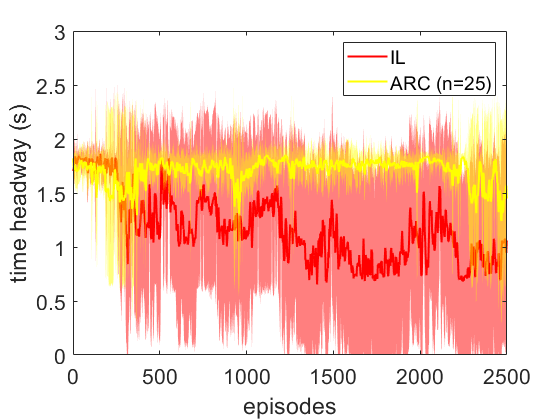}
	\caption{Minimum episode headway during Adversarial Testing. Averaged over 5 training runs, with standard deviation shown in the shaded region.}
	\label{fig_advtest}

\end{figure}

The full results of both tests are shown in Table \ref{tbl_simresults}. Firstly, we can see that the ARC with a fixed adversary and no distillation loss converges to an overly conservative driving behaviour, maintaining large distances from the vehicle in front, as shown by its average headway of 21s. Once the distillation loss is introduced, the ARC model with the fixed adversary is significantly less conservative with an average $t_h$ of 3.3s. However, this model significantly overfits to the adversary it is training against, and fails to generalise to naturalistic driving as well as against new adversaries. Once the adversary is trained simultaneously with the protagonist, we see the model generalise to different scenarios significantly better. The ARC ($n = 1$) model can now drive without collisions with an average headway of 2.02s in naturalistic driving, as well as showing improved robustness against new adversaries when compared to the fixed adversary model. However, it is worth noting that the model still shows greater vulnerability to new adversaries compared to the original IL policy. Once we utilise multiple parallel environments ($n > 1$) with different adversaries, we obtain improved robustness to new adversaries compared to the IL policy, while also demonstrating similar level of performance in naturalistic driving. As illustrated in Fig. \ref{fig_collsperadv}, the vulnerability of the ARC model to new adversaries reduces with increasing number $n$, up to 25. The minimum episode headway during the training of new adversaries for adversarial testing is illustrated in Fig. \ref{fig_advtest}, which shows the significant improvement in robustness with ARC. While it would be expected that the robustness of ARC increases further with the size of $n$, our results show that the best robustness is reached at $n = 25$. A potential reason for the lower robustness with $n = 50$, is that the global number of episodes for each ARC model was fixed at 2500. This means that as the number of environments increases, each environment collects less experience in total, and once the number of episodes per environment becomes too small there may not be enough experiences collected against the adversaries for the protagonist to learn how to counter them. This suggests there is a maximum number of environments that can be utilised for a given number of global training episodes. However, increasing the number of environments may still result in further improvement, if the number of global episodes is also increased.

The two testing frameworks demonstrate the benefit of the ARC approach. By starting with an initial policy susceptible to adversarial attack, and tuning it against adversarial policies, the policy becomes significantly more robust to such adversarial agents. Also, by utilising multiple environments in parallel, each using separate adversaries and training the policy asynchronously against all adversaries, the model gains superior generalisation and robustness. Furthermore, by utilising the distillation loss with knowledge from the IL network, the model avoids adapting overly conservative behaviour or overfitting to the adversarial scenarios, thereby ensuring the performance in the original training distribution is not degraded.

%%%%%%%%%%%%%%%%%%%%%%%%%%%%%%%%%%%%%%%%%%%%%%%%%%%%%%%%%%%%%%%%%%%%%%%%%%%%%%%%
\section{Conclusions}
In this paper, an approach to fine-tune the robustness and safety of a vehicle motion control policy was demonstrated. The approach was tested by fine-tuning an Imitation Learning control policy, which was shown to be vulnerable to adversarial agents. By training the IL policy against an ensemble of adversaries in multiple parallel simulations, it learned to counter the adversaries without overfitting to the behaviour of any single adversary. It was also demonstrated that after fine-tuning, the robustness to new adversaries is significantly improved, as demonstrated by the 90.25\% reduction in collisions when tested against new adversarial agents. Moreover, testing in natural driving scenarios demonstrated that by utilising a distillation loss, the performance in the policy's original training distribution is not compromised. Therefore, this work demonstrated a fine-tuning strategy, which uses adversarial learning to significantly improve model generalisation and robustness to out-of-distribution scenarios, without trading off performance in its training distribution.

This work opens up multiple potential avenues for future work. Investigating this fine-tuning strategy for different control policies or use-cases would be interesting. Moreover, identifying techniques which could limit the amount of training with a simulator in the loop could be useful for reducing the training times and increasing the flexibility of this framework. This could be done by either improving the sample efficiency of the adversarial reinforcement learning used in the ARC framework, or extending the framework such that some or all of the training can be done offline with no simulator (e.g. by using a dataset of interactions between the adversary and protagonist). More importantly, further testing of the adversarially robust control in real-world training environments would be useful to gain further insight on how this framework could be expanded for real-world autonomous vehicles. This work has demonstrated that the technique is effective in improving the driving policies' robustness when leveraging multiple simultaneous parallel simulations. To extend this in the real-world, one option would be to leverage multiple pairs of physical protagonist and adversarial agents, which then update a global network. Alternatively, sim-to-real transfer, an active area of research \cite{pan2017virtual, rusu2017sim, tobin2017domain, osinski2020simulation}, could be investigated to better leverage the faster training offered by simulators and minimising the amount of costly real-world training required.

\addtolength{\textheight}{-15cm}   % This command serves to balance the column lengths

%%%%%%%%%%%%%%%%%%%%%%%%%%%%%%%%%%%%%%%%%%%%%%%%%%%%%%%%%%%%%%%%%%%%%%%%%%%%%%%%

\section*{Acknowledgment}
This work was funded by the EPSRC under grant agreements (EP/R512217/1) and (EP/S016317/1).

%%%%%%%%%%%%%%%%%%%%%%%%%%%%%%%%%%%%%%%%%%%%%%%%%%%%%%%%%%%%%%%%%%%%%%%%%%%%%%%%

\bibliographystyle{IEEEtran}
\bibliography{ref_ganil}

\end{document}